\def\year{2017}\relax
\documentclass[letterpaper]{article}
\usepackage{aaai17}
\usepackage{times}
\usepackage{helvet}
\usepackage{courier}
\usepackage{graphicx}
\usepackage{color}
\usepackage{caption}
\usepackage{multicol,lipsum}
\usepackage{amssymb}
\usepackage{amsmath}
\usepackage{booktabs}
\usepackage{breqn}
\begin{document}

%
\title{An Artificial Agent for Robust Image Registration}
\author{Rui Liao, Shun Miao, Pierre de Tournemire, Sasa Grbic,\\ \textbf{\fontsize{12}{12}\selectfont Ali Kamen, Tommaso Mansi, Dorin Comaniciu}\\
Technology Center, Medical Imaging Technologies\\
Siemens Medical Solutions USA\\
Princeton, NJ 08540\\
}
\maketitle
\begin{abstract}
3-D image registration, which involves aligning two or more images, is a critical step in a variety of medical applications from diagnosis to therapy. Image registration is commonly performed by optimizing an image matching metric as a cost function. However, this task is challenging due to the non-convex nature of  the matching metric over the plausible registration parameter space and insufficient approaches for a robust optimization. As a result, current approaches are often customized to a specific problem and sensitive to image quality and artifacts. In this paper, we propose a completely different approach to image registration, inspired by how experts perform the task. We first cast the image registration problem as a "strategy learning" process, where the goal is to find the best sequence of motion actions (e.g. up, down, etc.) that yields image alignment. Within this approach, an artificial agent is learned, modeled using deep convolutional neural networks, with 3D raw image data as the input, and the next optimal action as  the output. To cope with the dimensionality of the problem, we propose a greedy supervised approach  for an end-to-end training, coupled with attention-driven hierarchical strategy. The resulting registration approach inherently encodes both a data-driven matching metric and an optimal registration strategy (policy). We demonstrate, on two 3-D/3-D medical image registration examples with drastically different nature of challenges, that the artificial agent outperforms several state-of-art registration methods by a large margin in terms of both accuracy and robustness. 
\end{abstract}

\section{Introduction}

The goal of 3-D medical image registration is to recover correspondences between two 3-D images acquired from 1) different patients, 2) the same patient at different time, and 3) different modalities e.g. Computed Tomography (CT), Magnetic Resonance Imaging (MRI), Positron Emission Tomography (PET) etc. The images are brought into the same coordinate system via various transformation models, e.g. rigid, affine, parametric splines, and dense motion fields \cite{Oliveira2014}. The aligned images could  then provide complimentary information for decision-making, enable longitudinal change analysis, or guide minimally invasive therapy \cite{James2013,Liao2013}. 

\begin{figure*}
    \centering
    \includegraphics[width=0.8\linewidth]{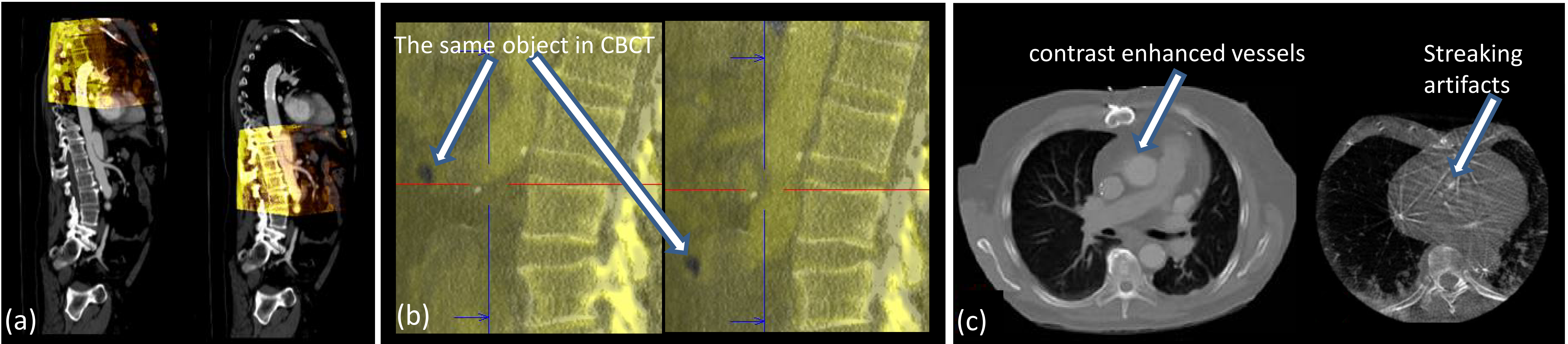}  
    \caption{(a)  Overlay of spine CT (grey) and CBCT (yellow) volumes before (left) and after (right) registration, with large differences in FOVs.  (b) Correct overlay of  CT and CBCT (left) versus wrong overlay by shift of one vertebra of CBCT (right). The shift is shown by the  movement of the dark object in CBCT, while the change in spine overlay is barely noticeable.  (c) Cardiac CT (left) and CBCT (right) volumes, with weak soft tissue contrasts and severe streaking artifacts in the CBCT.     }
    \label{fig:ExampleChallenges}
\end{figure*}

While medical image registration has been an active research area for more than two decades \cite{Markelj2010,Murphy2011}, fully automatic and robust 3-D registration  remains a challenging task that requires manual intervention for possible corrections. Up-to-date  image registration has been largely formulated as an optimization problem, where a generic matching metric is defined to measure the similarity of the image pairs to be registered \cite{Razlighi2013}. The transformation parameters between the image pairs are then estimated via maximization of the defined matching metric using an optimizer \cite{Rios2013}. This formulation faces two challenges. First, a generic matching metric is often non-convex over the plausible registration parameter space and generic optimizers perform poorly on non-convex problems. For example, the CT and cone-beam CT (CBCT) spine images in Figure~\ref{fig:ExampleChallenges}.a have very different field of views (FOVs), resulting in  local maxima due to the repetitive nature of vertebra (Figure \ref{fig:ExampleChallenges}.b). Second, a generic matching metric does not guarantee a good  alignment, e.g, when the data is noisy or with drastically different appearance due to different imaging physics. The cardiac case in Figure~\ref{fig:ExampleChallenges}.c shows contrast enhanced vessels in CT and severe streaking artifacts and weak soft tissue contrast in CBCT.

In this paper, we reformulate the registration problem by mimicking more closely how an expert performs image registration as a process of sequential actions of object recognition and manipulation. Motivated by recent advances in Deep Neural Networks (DNN) and Deep Reinforcement Learning (DRL) \cite{Mnih2015,Silver2016,Caicedo2015,Hausknecht2015,Neumann2016}, during training the agent learns a registration strategy (policy) via a DNN that maps the current state to the optimal next action that best improves the alignment. During testing, the agent applies the learned policy sequentially to align the images. As a result, the artificial agent inherently learns both a data-driven matching metric and a registration task-driven policy. 

Our main technical contributions are: 1) Instead of  one-shot regression mapping the raw image data to the registration parameters, which  is  often a very hard problem to learn, we decompose the registration task into a sequence of (often easier) classification problems, i.e. finding the best action among a limited set of possible solutions to improve the alignment. Repeating this process can result in a
converging solution.  2) We train the intelligent agent in a greedy supervised fashion, which is a magnitude more efficient with only a small fraction of the memory footprint than in standard DRL setup, where the agent learns through repeated trial and errors \cite{Mnih2015,Silver2016}; 3) we propose an effective data augmentation and sampling strategy so that the agent could be trained robustly using only a small number of labelled training pairs  available from patients; 4) For a combined robustness and accuracy, we propose a hierarchical registration framework relying on the trained networks from the coarse image layer to register successively the more refined (higher-resolution) image layers.

\section{Related Work }
\label{gen_inst}

{\bf 3-D Medical Image Registration} The most common strategy to reach robust intensity-based 2-D or 3-D image registration relies on multi-resolution strategy with local optimizers \cite{Thevenaz2000}. However, multi-resolution cannot cope with different FOVs or image artifacts. Global exhaustive search has been primarily used with 2-D image registration tasks due to its high computational complexity. Heuristic semi-global optimization schemes were proposed, e.g. simulated annealing \cite{Matsopoulos1999} and genetic algorithm \cite{Rouet2000}, however their computational cost for 3-D registration is still prohibitively high. Other more efficient global optimization techniques such as Bayesian optimization~\cite{Snoek2012} and trust region algorithms~\cite{Yuan2015} have been investigated for other applications but not yet widely adopted in medical image registration. Alternatively prior knowledge about the specific anatomies and medical workflow have been incorporated for specific registration tasks \cite{Lu2014,Miao2013}. Anatomical feature based 3-D registration were also proposed where landmarks \cite {Brounstein2011} or  surfaces  \cite{Chen2009} were extracted from the images and then matched. The accuracy  thus heavily depends on the segmentation methods used. More recently machine learning based hybrid method was proposed  \cite {Neumann2015}. However, the optimization process was still standard, and thus prone to local optimum.

{\bf Image Registration and Pose Estimation via DNN}   A DNN is an artificial neural network with multiple hidden layers of units between the input and output layers,  giving the potential of modeling complex data with fewer units than a similarly performing shallow network ~\cite {hinton2006reducing}. Convolutional neural network (CNN) is a feed-forward network in which the connectivity pattern between its neurons is inspired by the organization of the animal visual cortex and is observed to be most suitable for image processing tasks ~\cite {krizhevsky2012imagenet}. However while CNN has achieved state-of-the art performance in  image segmentation, image recognition, and image classification, there are only a few work addressing image registration using CNN. Unsupervised learning using CNN was proposed in \cite {Wu2015} to extract features for deformable registration. These features however were extracted separately from the image pairs and therefore may not  be optimal for registration purpose. A CNN-based regression approach was presented in \cite {Miao2016} to solve 2-D/3-D registration for device tracking from 2-D X-ray images. Optical flow estimation between 2-D RGB images has been proposed using CNN via supervised learning in \cite{Fischer2015}. A descriptor learned via CNN was described in \cite{Wohlhart2015} to encode both the identity and the pose of the 3-D object from 2-D RGB images. Their formulations however could not be directly applied to 3-D medical image registration where the displacement is large and there is no object model at hand. A learnable module, called spatial transformer network (STN), was introduced in \cite{jaderberg2015}. The focus of STN was not an accurate alignment of two images, but a rough transformation of a single input image to a canonical form, for the purpose of improved classification accuracy. Moreover its application has been demonstrated only on 2-D images, not  3-D volumes.      

{\bf Deep Reinforcement Learning} In Reinforcement Learning (RL), the agent learns  to perform certain tasks through a reward system, via successive trial and errors. While RL has been widely studied in game theory, control, operations research, robotics, etc., it is only with the recent breakthroughs in DRL, which combine RL with DNN, that it could be applied to more complex  problems, reaching human-level  performances (e.g. Atari game~\cite{Mnih2015} and Go~\cite {Silver2016}). In  \cite {Caicedo2015}  an active detection model for localizing objects in 2-D RGB images was trained using DRL. Similarly, an detection agent was trained using DRL\ for localizing landmarks in 3D CT images in ~\cite {ghesu2016artificial}. However, one of the main challenges in DRL is the training process, which can be extremely time-consuming. Guided policy search \cite{levine2013} and imitation learning \cite{kober2010} were proposed for more efficient RL via improved policy/data sampling, which however are not directly applicable to the end-to-end trainable DRL framework. In this paper, we  follow DRL framework but train the agent via greedy Deep Supervised Learning (DSL), and thus completely removes the need of an exploration history of the agent as in DRL, leading to drastically improved training efficiency.

\section{A Framework to Train An Intelligent Agent for Image Registration }
\label{headings}

\textbf{Problem Formulation}
Let $I_r:\mathbb{R}^3\mapsto\mathbb{R}$ be a reference image and $I_f:\mathbb{R}^3\mapsto\mathbb{R}$ be the floating image to be registered to $I_r$.
The goal of 3-D rigid-body image registration is to estimate the rigid-body transformation $T_g:\mathbb{R}^3\mapsto\mathbb{R}^3$ for which the transformed floating image $T_g \circ I_f$ is aligned with $I_r$ .
A 3-D rigid-body transformation $T$ can be represented by a column-wise $4\times 4$ homogeneous transformation matrix, such that a point $\vec p=[x,y,z,1]^\top$ (in homogeneous coordinates) is transformed to $T\vec p$.  Also $T$ can be parameterized by 6 parameters with 3 translations $[t_x, t_y, t_z]$ and 3 rotations $[\theta_x, \theta_y, \theta_z]$ as:
\begin{eqnarray}
T(t_x, t_y, t_z, \theta_x, \theta_y, \theta_z) = \begin{bmatrix}
1 & 0 & 0 & t_{x}\\
0 & c\theta_{x} & -s\theta_{x} & t_{y}\\
0 & s\theta_{x} & c\theta_{x} & t_{z}\\
0 & 0 & 0 & 1\\
\end{bmatrix} \nonumber \\ \times
\begin{bmatrix}
c\theta_{y} & 0 & s\theta_{y} & 0\\
0 & 0 & 0 & 0\\
-s\theta_{y} & 0 & c\theta_{y} & 0\\
0 & 0 & 0 & 1\\
\end{bmatrix} \times
\begin{bmatrix}
c\theta_{z} & -s\theta_{z} & 0 & 0\\
s\theta_{z} & c\theta_{z} & 0 & 0\\
0 & 0 & 1 & 0\\
0 & 0 & 0 & 1\\
\end{bmatrix}\
\label{equ:transformation} 
\end{eqnarray}
where $c$ denotes $\cos$ and $s$ denotes $\sin$.  

We cast the problem of finding $T_g$ as a Markov Decision Process (MDP)~\cite {Bellman1957} defined by the tuple $\{S, A, \tau, r, \gamma\}$, where $S$ are the set of states in which an artificial agent can be, $A$ is the set of actions it can take,  $\tau$ is the stochastic transition function associated to a state-action-state relationship, $r$ is the reward the agent receives when taking a specific action at a specific state, and $\gamma$ is the discount factor that controls the importance of future rewards (equal to 0.9 in this work to favor long term rewards).
In our system, at each time point $t$, the state $s_t$ is defined by the current transformation $T_t$, and $T_g \circ T_t^{-1}$could be parameterized as $v_t\in{R}^6$ using Eqn~\ref{equ:transformation}. The associated observation of the system at time $t$ is the difference image $d_{t} = I_{r} - T_t \circ I_f$.
Then, the agent chooses an action $a_t \in A$ to alter the state  to improve  image alignment by $T_{t+1} = a_t \circ T_t$ . 
The action set $A$ consists of 12 candidate transformations that lead to the change of $\pm1$ in one element of $v_{t+1}$ compared to $v_t$ (i.e. $\pm 1$mm for translations or $\pm 1^\circ$ for rotations).
 The transition function $\tau$ is defined by giving equal weights to all available actions. The reward function $r$ is described in more details below. During training, the agent learns a registration policy (i.e. a strategy of sequential actions) that maps the current state $s_t$ to the optimal action $a_{t}^{*}$, defined as the action that best improves the alignment. During testing, the agent applies the learned policy in a sequence of $N$ consecutive actions, $ \{a_{1}^{*}, \dots, a_{N}^{*}\}$ to approach the correct alignment. 

\begin{figure}
    \centering
    \includegraphics[width=\linewidth]{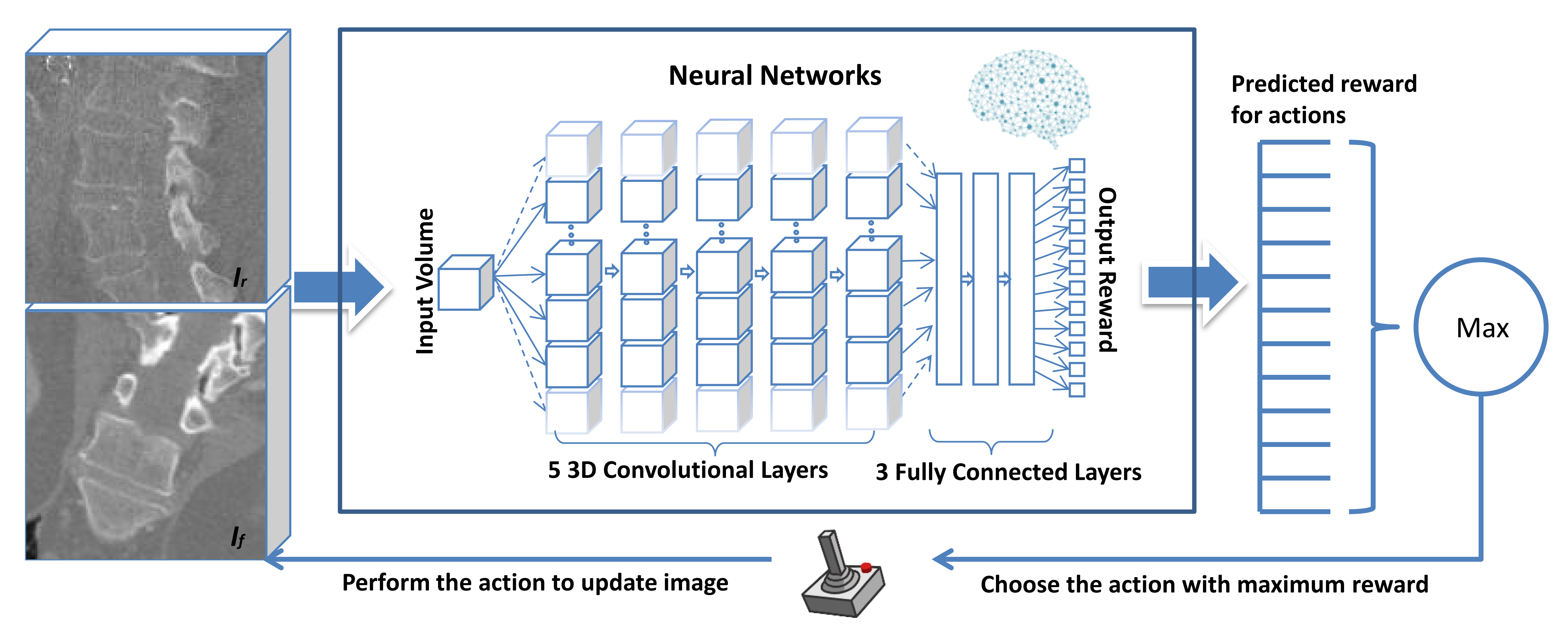}  
    \caption{A framework for the intelligent agent to perform registration. }
    \label{fig:infrastructure}
\end{figure}

\textbf{Target $Q$-Learning with A Supervised Registration Path} The core problem is to find a policy that guides the decision process of the artificial agent. In \cite{Mnih2015,Silver2016}, this policy learning process is formulated as a RL problem, where the optimal action-value function $Q_{t} (s_{t},a_{t})= \max_{\tau}\mathbb{E}[r_{t}+\gamma r_{t+1}+\gamma^2 r_{t+2}+...|s_{t},a_{t},\tau]$   is approximated by a DNN and learned following the Bellman equation as an iterative update. However, unguided exploration of the agent and iterative update of $Q$ can result in a low training efficiency, as the agent has to try many combinations before reaching an effective policy. Instead, we propose to supervise the training by instructing the agent to follow a greedy registration path, mimicking how human register two objects in a most efficient manner.
Specifically, the ``optimal'' action $a_{t}^{*}$ along the supervised registration path is defined as the action that minimizes the ``distance'' between the new transformation $a_t \circ T_t$ and the ground truth transformation $T_g$:  
\begin{equation}
a_{t}^{*} = \min _{a_{t\in A}} D( T_g, a_{t} \circ T_t ).
\label{equ:optimalaction}
\end{equation}
where $D( T_g, T )$, the distance between two transformations $T$ and $T_g$, is defined as the L2 norm of the 6-D parameters of $T_g \circ T^{-1}$, using the parameterization described in Eqn~\ref{equ:transformation}.
If more than one actions lead to the same minimal distance, any of these actions could be taken with equal probability.  Without loss of generality, in this paper the agent is allowed to explore the transformation parameter space only within $\pm 30$mm  for $t_{x},t_{y}, \pm 150$mm  for $t_{z}$, and $\pm 30^{\circ}$ for $\theta_{x},\theta_{y},\theta_{z}$, corresponding to the maximum possible mis-alignment of the two volumes to be registered (more details are provided in the Experiments section). 

In this setup, the action-value $Q$-function can be calculated explicitly via a recursive function, assuming  the agent is allowed to run sufficient number of steps to reach the correct alignment following the supervised greedy path:
\begin{dmath}
Q(s_{t},a_{t} ) = \begin{cases}r(s_{t},a_{t} )_{}+\gamma Q(s_{t+1},a_{t+1}^{*} )\: & \textrm{if} \: D(T_g, a_t \circ T_t)>\epsilon\ \\
r(s_{t},a_{t} )+R & \textrm{o.w.} \\
\end{cases}
\label{equ:Qfunction}  
\end{dmath}
where the immediate reward $r(s_{t},a_{t})$ for action $a_{t}$ is:
\begin{equation}
r(s_{t},a_{t} ) =  D( T_g, T_t ) - D( T_g, a_{t} \circ T_t ).\\ 
\label{equ:reward} 
\end{equation}
The agent is considered to reach the correct transformation  when its distance to $T_{g}$ is within the tolerance $\epsilon = 0.5$. When this happens, the agent receives a bonus reward $R=10$. Interestingly it can be shown (in Appendix A) that if the agent is allowed to take continuous actions in the 6-D transformation parameter space with step size 1, i.e.  the only constraint is $\Vert v_{t+1}-v_t\Vert_2=1$, then $Q(s_{t},a_{t}^{*})$ calculated by Equation \ref{equ:Qfunction} is the maximum of the action-value function  $Q_{t} (s_{t},a_{t})$ with respect to action $a_t$. This means the trained agent can perform registration by simply choosing the action with the largest $Q$ in the testing phase.   

Following \cite{Mnih2015,Silver2016} we use deep CNN to represent $Q$ in Equation~\ref{equ:Qfunction}. The input to the network is the current difference image $d_t$,  the output of the network has 12 nodes, where each corresponds to one of the 12 actions in the action set $A$, and the loss function is: 
\begin{equation}
Loss = \sum _{k=1}^M\sum _{a_{i=1...12}\in A,}\Vert y_{i} (d_{k})- Q(s_{k},a_{i} )\Vert_{2} 
\end{equation}
where $y_{i} (d_{k})$ is the $i$-th $(i=1...12)$ output of the CNN for the $k$-th sample among \textit{M} training samples. Our CNN training scheme, called Deep Supervise Learning (DSL), has two major advantages  compared to DRL. First, our target Q-function is given analytically without iterative estimation so that the network could be trained much more efficiently and with a more stable convergence property. Second, our target Q calculation does not require the exploration history of the agent, meaning that we could sample the data randomly with little correlations and thus reduce memory requirements. Both advantages are critical to make 3-D registration learning possible with large 3-D volumes as inputs.      
 
\textbf{Hierarchical Image Registration}
Since our inputs to the network are two large 3-D medical images, which can be up to $512\times512\times512$ voxels for instance, the size of the input is of critical importance for practical use. For a combined robustness and accuracy, we propose a hierarchical strategy based on attention. The idea is to train two separate CNNs, both using 64$\times$64$\times$64 volumes as the input but with different resolutions and FOVs. The first CNN is trained for coarse alignment  using down-sampled volumes with a lower resolution but larger FOV, helping the agent to gain global anatomical understanding and thus able to perform robust alignment of the object without being trapped into local optimum even when the initial displacement is large. The second CNN uses a high-resolution volume with a limited FOV and focuses on aligning the object as accurately as possible despite the limited FOV. The registration task is then performed as follows. First, the  agent applies the first  CNN to roughly align the object using $N_{1}$ (empirically set to 200) sequential actions. Then, following a similar approach as in~\cite{Simonyan2013}, we use single back-propagation pass to  compute the derivative of the sum of the outputs of the first CNN  with respect to the input image to get a saliency map $\Omega$. $\Omega$ determines the importance of a given pixel in influencing the outcome of the CNN network for the first step of coarse registration, and those most influencing pixels (presumably corresponding to the spine) are selected via thresholding using 95th percentile, and their geometrical mean weighted by their importance is calculated as the center of the region of interest (marked by the blue rectangle box in Figure \ref{fig:SaliencyMap}) for the second step of refined registration. Finally, the region of interest is extracted from the high-resolution volume, and starting from the final position obtained in the first step, the agent applies the second CNN with $N_{2}$ (empirically set to100) sequential actions. 

\textbf{Data Augmentation and Sampling Strategy}
Training the CNN requires labeled training pairs with known transformations $T_{g}$. Unfortunately, such ground truth is  not easily obtainable in the medical domain. We thus propose to augment the available labeled data in two ways. First, each aligned pairs are artificially de-aligned using randomly generated rigid-body motions. Denser sampling at the transformation parameter space close to the ground truth transformation $T_g$ is also performed for finer training of the network close to the solution. Second, each aligned pairs are geometrically co-deformed by affine transformations $T_A$, where $I$ is the 4x4 identity matrix and all the elements in $[c_{ij}]_{i=1,2,3, j=1,2,3}$ for shearing are independently and randomly generated from [-0.25, 0.25], to cover possible anatomical variations among patients in  sizes and shapes:
\begin{equation}
T_{A}=I+\begin{bmatrix}
c_{11} & c_{12} & c_{13} & 0\\
c_{21} & c_{22} & c_{23} & 0\\
c_{31} & c_{32} & c_{33} & 0\\
0 & 0 & 0 & 1\\
\end{bmatrix}
\label{equ:affineTransform}  
\end{equation}
  
\section{Experiments}
\label{others}

\textbf{Experiment Setup}
We experimented  on two 3-D medical image registration data sets. \textbf{E1:} Abdominal spine CT and CBCT, where the main challenging is that  CT has a much larger FOV than CBCT, leading to many local optima in the
registration space due to the repetitive nature of the spine. Indeed offset by one vertebra could be relatively unnoticeable even for human eyes (Figure \ref{fig:ExampleChallenges}.b). Registration accuracy was measured by 3-D target registration error (TRE), defined as the average 3-D Euclidean distance between the transformed landmarks and the corresponding ground truth  for 32 landmarks on the edge of the vertebrae.  Success rate was evaluated by TRE$\le$10 mm \cite {Miao2013}. \textbf{E2:} Cardiac CT and CBCT, where the main challenge is the poor quality of CBCT  with severe streaking artifacts and weak soft tissue contrast at the boundary of the object to be registered, i.e. the epicardium. Registration accuracy was measured by  mesh-to-mesh error (MME). Point-to-triangle distances were calculated for all  the vertices of the segmented epicardium meshes and then averaged to get the final  MME. Success rate was evaluated by MME$\le$20 mm \cite {Neumann2015}.   Spine landmarks and epicardium segmentation were performed by experts.  Iterative closest point registration~\cite {Besl1992} followed by visual inspection and manual editing whenever necessary was performed to provide the ground truth alignment. 

  The testing  and training data were blindly separated and  guaranteed to be from different patients to minimize their correlations.
 Cross-validations were furthermore performed by 5  different blind data-splits for both E1 and E2 (validation for one data-split took  4 days on a 24 core + GeForce TitanX computer for data augmentation and training). For each data-split, there were   82 pairs for training and 5 pairs for testing for E1, and 92 pairs for training and 5 pairs for testing for E2. Since the goal for registration is to estimate the transformation between images, multiple test cases could be generated from one test pair by perturbing their initial alignment. Specifically, each test pair was randomly de-aligned 10 times using rigid-body perturbation  within the same range as those used for generating the corresponding training data, resulting in 5x10x5x2=500 test cases. Each of the 500 test cases is unique and sufficiently different from training data (details are given in the next section) because: 1) human anatomy naturally varies; 2) registration results of medical images vary a lot for different initial alignment due to its highly non-convex nature; 3) $60^{12}$ samples are needed to fully cover this 6x2=12D parameter space spanned by perturbations of $\pm$30x30x30 degrees and $\pm$30x30x30 mms for both reference and floating images, and thus 5M training data samples only an extremely small portion $(\approx  1/5*10^{14})$ of this space.

\textbf{Network Architecture and Training Details} We used the same network architecture and meta-parameters for both applications and both coarse and fine registration. The network consists of 5 convolutional layers followed by 3 fully connected layers. The convolutional layers use 8, 32, 32, 128, 128 filters, all with 3x3x3 kernels. The first 2 convolutional layers are each followed by a 2x2x2 max-pooling layer. The 3 fully-connected layers have 512, 512, 64 activation neurons, and the output has 12 nodes corresponding to the 12 possible actions in $A$. Each layer is followed by a nonlinear rectified layer, and batch normalization is applied to each layer. During training, each training pair was augmented 64000 times, leading to more than 5M training data for each data-split. To train the CNN for coarse registration, rigid-body perturbation was randomly generated within $[\pm30\text{mm},\pm30\text{mm}, \pm30\text{mm}, \pm 30^{\circ},\pm 30^{\circ},\pm 30^{\circ}]$ for E2, and $[\pm 30\text{mm}$, $\pm 30\text{mm}, \pm 150\text{mm}, \pm 30^{\circ},\pm 30^{\circ},\pm 30^{\circ}]$ for E1 to cover the large FOV in the head-foot direction in spine CT.
To train the CNN for refinement registration, rigid-body perturbation range was reduced to $[\pm 5\text{mm}, \pm 5\text{mm}, \pm 5\text{mm}, \pm 5^{\circ},\pm 5^{\circ},\pm 5^{\circ}]$.
We used RMSprop update without momentum and  a batch size of 32. The learning rate was 0.00006 with a decay of 0.7 every 10000 mini-batch based back-propagations.

\subsection{Comparison between DSL and DRL}

\begin{figure}
        \centering
        \includegraphics[width=0.32\linewidth]{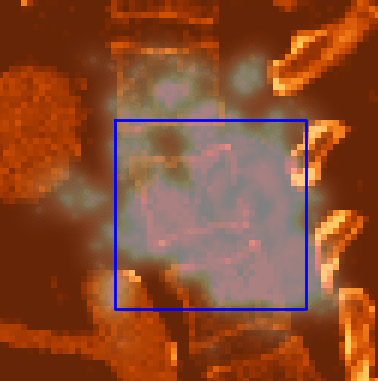}
        \includegraphics[width=0.32\linewidth]{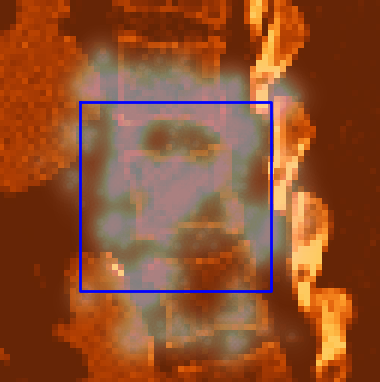}
        \includegraphics[width=0.32\linewidth]{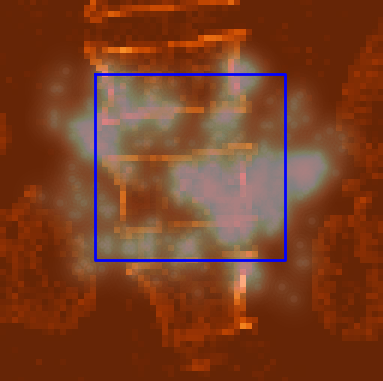}
        \captionof{figure}{Examples of saliency maps and attention of focus.}
        \label{fig:SaliencyMap}
\end{figure}

\begin{figure}
        \centering
        \includegraphics[width=0.95\linewidth]{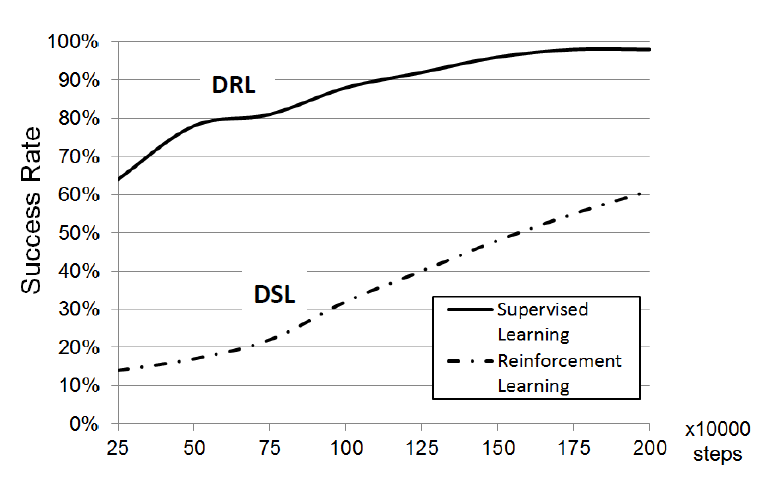}
        \captionof{figure}{Plot of success rates versus the number of training steps for DSL and DRL on 100  test samples.}
        \label{fig:SLvsRL}
\end{figure}

We evaluated the  efficiency of our proposed DSL in comparison with DRL~\cite{Mnih2015} on a modified 2-D registration problem using the spine data as a demonstration (3-D registration could not be learned using DRL due to the prohibitive memory requirement of memory replay, i.e. $\sim$2T for  64x64x64 volumes and replay memory history of 1 million). In particular, 2000 2-D MPR\ image pairs were extracted from 82 aligned CT and CBCT spine pairs using various Multiplanar Reconstruction (MPR) cuttings. For DSL, these 2-D images were  artificially de-aligned by a random perturbation within $[\pm30\text{mm},\pm30\text{mm}, \pm 30^{\circ}]$ to generate 2M image pairs for training. For DRL, 2M exploration steps were performed by the agent using the immediate reward in Equation~\ref{equ:reward}. The network architecture was modified slightly to take 128x128 2-D images as the input, and the output has 6 nodes corresponding to  6 possible actions in changing $[t_x,t_y,\theta]$. The network architecture and training meta-parameters were the same for DSL and DRL. It is clear from Figure~\ref{fig:SLvsRL}. that the proposed DSL was much more efficient than DRL and achieved significantly better results when the same number of  training steps were performed using the same training time (i.e. 1 day). 

\begin{table*}
  \caption{Comparison of Registration Results (\#1 and \#2 results are marked in red and blue).}
  \label{table:registrationresults}
  \centering
  \begin{tabular}{lcccccccc}
    \toprule
    & \multicolumn{4}{c}{\textbf{Spine (E1)} (TRE mm)} & \multicolumn{4}{c}{\textbf{Heart} \textbf{(E2)} (MME mm)}
    \\
    \cmidrule(r){2-5}
    \cmidrule(r){6-9}                 
    \textbf{Methods} &\textbf{Success}\ &\textbf{10th }&\textbf{50th} &\textbf{90th} &\textbf{Success}\ &\textbf{10th}       &\textbf{50th} &\textbf{90th}\\
    Ground Truth &N/A &0.8  &0.9  &1.2  &N/A\ &2.1  &4.0  &5.9 \\
    Initial Position &N/A &35.5  &73.9 & 116.2 & N/A   &9.2  &22.8 &30.5\\
    ITK\cite{Ibanez2005} & 12\% & 1.9 & 77.3 &130.4 &14\% &14.9 &34.9  &47.6\\
    Quasi-global\cite{Miao2013}\ & 20\% & \color{blue}{1.6} & 60.9& 136.2 & 14\% &16.2 &35.9  &58.7\\
    Semantic registration\cite{Neumann2015} & 24\%\ &3.0  &34.9 &71.0 &72\%\ &7.6 &15.3  &30.6\\
    \textbf{Proposed method} &\color{red}{92\%} &1.7 &\color{blue}{2.5} &\color{red}{3.8} &\color{red}{100\%} &\color{red}{3.2} &\color{red}{4.8}  &\color{red}{6.9} \\
    Human registration & \color{blue}{70\%} & \color{red}{0.8} &\color{red}1.6 & \color{blue}15.8& \color{blue}96\%&\color{blue}4.0 &\color{blue}6.2  &\color{blue}13.4   \\
    \bottomrule
  \end{tabular}
\end{table*}

\begin{figure*}
    \centering
    \includegraphics[width=380pt]{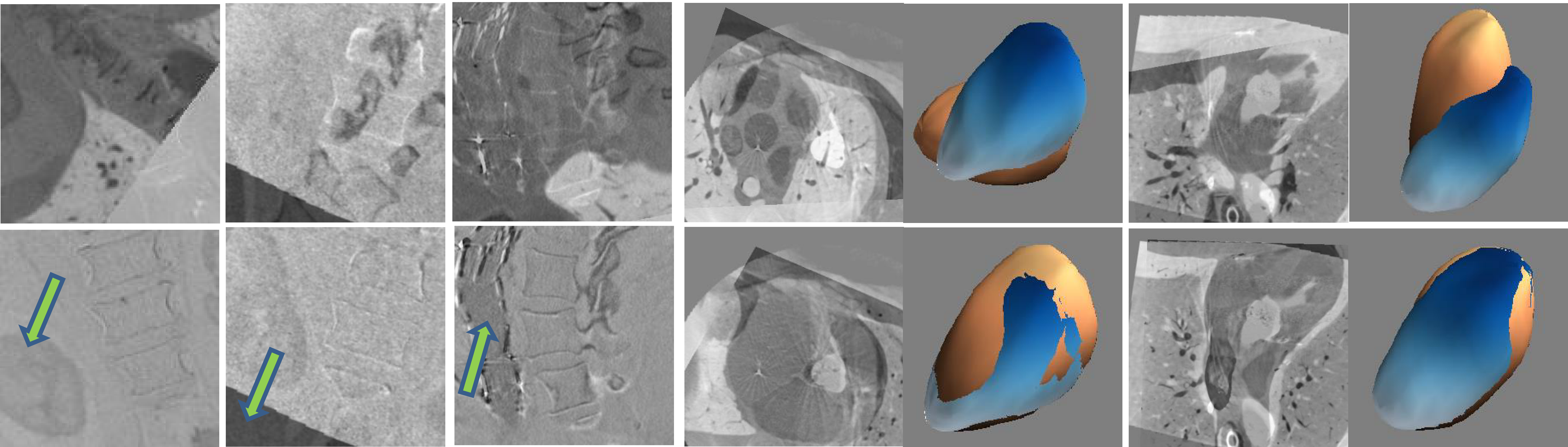}  
    \caption{Registration examples  shown as the difference  between the reference and floating images, before (upper row) and after (lower row) registration. The mesh overlay before and after registration is shown for the epicardium use case (E2) for improved visualization.  }
    \label{fig:RegistrationExamples}
\end{figure*}

\subsection{Evaluation of the Proposed 3-D Image Registration}
\textbf{Baseline Methods and Human Registration}
Our framework was quantitatively compared with three state-of-the-art 3-D image registration methods as well as human manual registration. \textbf{M1}: \textit{\textbf{ITK registration}} \cite {Ibanez2005}, a popular open source medical imaging library, where Mutual
Information (MI) computed based on 50 bins for
the histogram as proposed by \cite {Mattes2001} were used for the matching metric, and optimization was obtained using multi-resolution optimizer based on a variant of gradient descent for rigid versor transformations.  \textbf{M2}: \textbf{\textit{Quasi-global search }}\cite {Miao2013},  where 2-D anatomy targeted projections were generated to surrogate the original volume,  allowing for a large number of matching metric evaluations, approximating global-search. Implementation details followed \cite{Miao2013}. \textbf{M3}: \textbf{\textit{Semantic registration}}  \cite {Neumann2015}, where spine or epicardium was extracted from  CT volumes and a probabilistic map was calculated from CBCT volumes using probability boosting tree (PBT) \cite {Neumann2015}, and were then aligned via iterative optimization.
600  CT\ and 393  CBCT\ volumes were used for training for epicardium detection, and 82  CT and 82  CBCT volumes were used for training for spine detection. Implementation details followed \cite {Neumann2015}. \textbf{M4}: \textbf{\textit{Human registration}} by 5 different users. A tool allowing 6 degree of freedom manipulation of the volume with 3 MPR views and one volume view for the overlays was used. Two manual annotations were performed for each test pair using different initial perturbations by each user. 

\textbf{Evaluation of the Proposed Method}
Hierachical registration was applied for E1, and its effectiveness is demonstrated in Figure \ref{fig:SaliencyMap}. The median error was reduced from 3.4mm after applying the first CNN to 2.5mm after applying the second CNN. For {E2}, the MME was noticeable even for ground truth transformation due to the large, non-rigid deformation between CT and CBCT. Therefore the refinement step was not necessary and was not applied.
Quantitative results are summarized in Table \ref{table:registrationresults}. It is clear that the agent could perform reliable 3-D registration and even surpass human  performance when the cases were extremely challenging. Specifically, for E1, the agent could reliably overcome local maxima and was not confused by the highly similar appearance of the neighboring vertebrae. Furthermore, the agent was robust to interfering objects and artifacts, as highlighted by the green arrows in Figure~\ref{fig:RegistrationExamples} (from left to right: kidney, black background outside the image, and the deployed stent grafts). For E2, the weak soft tissue contrast and severe streaking artifacts makes reliable registration extremely challenging, even for human eyes. The  agent, however, was able to  learn the registration cues from raw high-dimensional training data, despite the low signal-to-noise ratio of the object to be registered. The results demonstrate that while the action of the agent  is limited to a set of local movements for each step, thus making the training of the network easier compared to one-shot decision (regression), the contextual understanding and overall strategy of the agent is indeed global, helping the agent avoid local optimum and achieve robust registration.

\textbf{Comparison with State-of-the-Art Methods}
Contrary to the proposed method, M1 and M2 easily failed in the challenging cases, leading to relatively low success rates. It should be noted that M2 performed much worse compared to the values reported in~\cite{Miao2013} due to the large rotations in our data, invalidating the assumptions made in~\cite{Miao2013}. While M3 performed more reliably compared to M1 and M2, it required a significantly larger number of training examples than our agent, and the performance deteriorated significantly when the number of training samples was limited as in  E1. The limitation comes from the fact that M3 does not inherently treat image registration as a problem of establishing the correspondence, but rather segments the objects  from the two volumes separately, followed by a standard iterative optimization scheme that is prone to local optimum. 

\section{Discussions and Conclusion}
This paper presents a novel 3-D rigid-body registration method based on artificial intelligence, where an agent is trained end-to-end to perform the registration task. The proposed framework is generic and the same network  hyper-parameters were used for all the experiments (most parameters were determined by simply adopting the values popularly used in the literature such as the discount factor $\gamma$, and some parameters were chosen given certain constraints, such as the bonus reward  $R>\frac{\gamma}{1-\gamma}$ ). In addition the training scheme is very efficient and requires only a relatively small number of labeled data.  We demonstrated on challenging cases that our  agent can outperform other state-of-the-arts methods by a large margin and even beat human performance when the difference in object appearance is subtle. This significantly superior performance on multiple applications without any hand-engineering in the training pipeline indicates that the propose method could potentially bring a new paradigm in medical image registration, a very challenging problem in practice. 

While there is no theoretical guarantee that the agent could finally achieve correct registration, in practice the agent is never observed to produce (large) cyclical movements but always converges to one position (correct or wrong). This is presumably due to the fact that our supervised registration path is (approximately) a straight line (thus far from a circle) in the registration parameter space. Randomization is also introduced by allowing the agent to take the best three actions with a given probability. The next step is to train an additional action of "stopping" so that the agent could stop early when correct registration is achieved. Heuristically a denser sampling at the transformation parameter space close to the ground truth transformation is performed for finer training of the network because the observation-action mapping close to the solution is more complicated compared to other regions. There could be other regions starting from where correct registration is more difficult to reach (accounting for 8\%\ failed cases in spine). A possible enhancement is to use more samples from those regions for training (i.e. boosting). 

Since the proposed DSL\ framework has exactly the same network input and output as the DRL framework, combination of our DSL  with DRL is straightforward, e.g. using DRL to refine the policy learned by DSL, which potentially allows the agent to learn a registration path that is more suitable  than the most greedy one for some tricky registration tasks. This combination is currently under investigation. Other future works include in-depth analysis of the trained networks, further evaluation on other use cases, and extension to higher dimensionality registration problems.  

\textbf{Disclaimer:} This feature is based on research, and is not commercially available. Due to regulatory reasons its future availability cannot be guaranteed.

\subsubsection*{Appendix A}
        \textbf{Proposition.} $Q(s_{t},a_{t}^{*})$ (calculated in Equation 3) for the optimal action $a_{t}^{*}$ (defined in Equation 2) is the maximum of the action-value function  $Q_{t} (s_{t},a_{t})$ with respect to action $a_t$, given the agent is allowed to take continuous actions with step size 1 in the 6-D transformation parameter space (i.e. the only constraint on the action is that $\Vert v_{t+1}-v_t\Vert_2=1$), and the agent receives a bonus $R>\frac{\gamma}{1-\gamma}$ when reaching the ground truth transformation (i.e.  $D(T_g, a_t \circ T_{t})<0.5$). 
        
        \textbf{Lemma 1.}  For all the continuous actions with step size 1, the maximum immediate reward $r$ defined in Equation 4 is 1.

        \textit{Proof.} $r(s_{t},a_{t} ) =  D( T_g, T_t ) - D( T_g, a_{t}  \circ T_t ) = \Vert v_{t}\Vert-\Vert v_{t+1}\Vert_2\leqslant \Vert v_{t}-v_{t+1}\Vert_2=\Vert v_{t+1}-v_t\Vert_2=1$.

        \textbf{Lemma 2.}   $Q(s_{t},a_{t}^{*} )\rightarrow\frac{1}{1-\gamma}$ as ${D(T_g, T_{t}) \rightarrow }\infty$, with a monotonic decrease.
        
        \textit{Proof.} Assume it takes the agent \textit{p+1}  steps from the current position  to reach the correct transformation using the optimal actions with step size 1, then $Q(s_{t},a_{t}^{*} )=F(p+1)=\sum _{i=0...p}\gamma^{ l}  +\gamma^{ p}R=\frac{1-\gamma^{ p+1}}{1-\gamma} +\gamma^{ p}R>\frac{1-\gamma^{ p+1}}{1-\gamma} +\frac{\gamma^{ p+1}}{1-\gamma}=\frac{1}{1-\gamma}$, and $ F(p+1_{})-F(p)=\gamma^{ p}+(\gamma^{ p}-\gamma^{ p-1})R<\gamma^{ p} +(\gamma^{ p}-\gamma^{ p-1})\frac{\gamma}{1-\gamma}=0,\forall p>0$. In addition, by definition $\gamma<1$,  so $F(p+1)=\frac{1-\gamma^{ p+1}}{1-\gamma} +\gamma^{ p}R\rightarrow\frac{1}{1-\gamma}$ as $p\rightarrow\infty$. 
        
        \textbf{Lemma 3.}   $Q(s_{t},a_{t}^{*}) \geqslant Q(s_{t},a_{t} ),\forall a_{t} \in A $.
        
        \textit{Proof.}
        Since $D(T_g, a_t^* \circ T_{t}) \leq D(T_g, a_t \circ T_{t})$, we get $r(s_{t},a_{t}^{*} )\geqslant r(s_{t},a_{t} )$ (Equation 4)
        and
        $Q(s_{t+1}^{*},a_{t+1}^{*})\geqslant Q(s_{t+1},a_{t+1}^{*})$, where  $s_{t+1}^{*}$ is the resulting state at $t+1$ by taking the optimal action $a_t^*$ at time $t$ (Lemma 2). Therefore $Q(s_{t},a_{t}^{*} ) =r(s_{t},a_{t}^{*} )+\gamma Q(s_{t+1}^{*},a_{t+1}^{*})\geqslant Q(s_{t},a_{t} )=r(s_{t},a_{t} )+\gamma Q(s_{t+1},a_{t+1}^{*})$.

\bigskip
\bibliographystyle{aaai}
\bibliography{refs}

\end{document}